\documentclass[11pt]{article}

\usepackage[final]{acl}

\usepackage{times}
\usepackage{latexsym}

\usepackage[T1]{fontenc}

\usepackage[utf8]{inputenc}

\usepackage{microtype}

\usepackage{inconsolata}

\usepackage{graphicx}

\usepackage{amsmath}
\usepackage{multirow}
\usepackage{multicol}
\usepackage{hhline}

\usepackage{booktabs}
\usepackage{multirow}
\usepackage{siunitx}

\usepackage{xcolor}
\newcommand{\best}[1]{\textbf{#1}}
\newcommand{\sbest}[1]{\underline{#1}}

\title{Pretraining and Benchmarking Modern Encoders for Latvian}

\author{Arturs Znotins \\
  Institute of Mathematics and Computer Science, University of Latvia\\
  29 Raina bulv., Riga, LV-1459, Latvia\\
  \texttt{arturs.znotins@lumii.lv} \\}

\begin{document}
\maketitle
\begin{abstract}
Encoder-only transformers remain essential for practical NLP tasks. While recent advances in multilingual models have improved cross-lingual capabilities, low-resource languages such as Latvian remain underrepresented in pretraining corpora, and few monolingual Latvian encoders currently exist. We address this gap by pretraining a suite of Latvian-specific encoders based on RoBERTa, DeBERTaV3, and ModernBERT architectures, including long-context variants, and evaluating them across a diverse set of Latvian diagnostic and linguistic benchmarks. Our models are competitive with existing monolingual and multilingual encoders while benefiting from recent architectural and efficiency advances. Our best model, lv-deberta-base (111M parameters), achieves the strongest overall performance, outperforming larger multilingual baselines and prior Latvian-specific encoders. We release all pretrained models and evaluation resources to support further research and practical applications in Latvian NLP.

\end{abstract}

\section{Introduction}
Large language models have shifted much of NLP research toward decoder-only architectures. However, encoder-only Transformers remain fundamental to practical natural language processing. They provide strong contextual representations for classification, sequence labeling, and span-based prediction, where systems require reliable token- and sentence-level outputs rather than free-form generation \cite{weller2025seqvsseqopen}. Furthermore, encoders are a key component of modern retrieval-augmented generation (RAG) pipelines, serving as the backbone for retrievers and embedding models. Encoder models are also typically more compute- and memory-efficient than autoregressive LLMs, making them well suited for high-throughput and latency-sensitive use-cases.

While multilingual encoders like mBERT and XLM-R enable robust cross-lingual transfer, scaling to a diverse set of languages introduces a capacity-–coverage trade-off \cite{conneau-etal-2020-unsupervised}. Because parameters and vocabularies are shared across languages with highly imbalanced pretraining corpora, low-resource languages often suffer from diluted capacity and insufficient training signals, leading to suboptimal representations.
Monolingual pretraining enables better language-specific tokenization and, when sufficient unlabeled data is available, often yields stronger downstream performance \cite{wang-etal-2020-negative, rust-etal-2021-good}. Therefore, monolingual encoders can provide a more efficient route to high-quality representations for a target language than relying on a shared multilingual model.

An open question in encoder design concerns the trade-offs among recent architectural choices. ModernBERT \cite{warner-etal-2025-smarter} emphasizes throughput and long-context support with modern attention implementations, whereas DeBERTaV3 \cite{he2023debertav3improvingdebertausing} utilizes disentangled attention and replaced-token detection (RTD). However, it remains unclear how these design differences translate to low-resource, monolingual settings.

In this work, we introduce a suite of Latvian pretrained encoder models and a unified evaluation benchmark. Using a large-scale Latvian corpus, we pretrain RoBERTa-, DeBERTaV3-, and ModernBERT-based encoders under comparable recipes, including long-context variants up to 8{,}192 tokens, and evaluate them across a diverse set of Latvian diagnostic and linguistic benchmarks.
Our best lv-deberta-base model achieves the highest overall score, substantially outperforming multilingual baselines and previously released Latvian models.

Our contributions are as follows:
\begin{itemize}
    \item We release a suite of Latvian pretrained encoders (RoBERTa, DeBERTaV3, ModernBERT) across multiple sizes, including long-context variants\footnote{\url{https://huggingface.co/collections/AiLab-IMCS-UL/latvian-text-encoders}}.
    \item We introduce a unified benchmark suite for evaluating Latvian pretrained encoder models\footnote{\url{https://github.com/LUMII-AILab/latvian-encoders}}.
\end{itemize}

\section{Related Work}
Since the release of BERT \cite{devlin-etal-2019-bert}, encoder models have evolved through improvements in both architecture and training objectives. RoBERTa demonstrated the importance of optimized training and data scaling by showing that BERT's performance can be significantly improved through longer training, more data, and refined objectives, without architectural changes \cite{liu2019roberta}. DeBERTa introduced disentangled attention mechanisms that separately model token content and relative position \cite{he2021deberta}. DeBERTaV3 further improved sample efficiency by combining disentangled attention with ELECTRA-style replaced-token detection objectives, improving sample efficiency \cite{he2023debertav3improvingdebertausing}. More recently, ModernBERT \cite{warner-etal-2025-smarter} and NeoBERT \cite{breton2025neobert} revisit encoder design with modern attention implementations and efficiency-oriented choices to enable high throughput and long-context processing.

Latvian is supported in multilingual encoders such as mBERT \cite{devlin-etal-2019-bert}, where it is limited to the Latvian Wikipedia subset (\textasciitilde{}25M tokens), and in larger-scale models such as XLM-R \cite{conneau-etal-2020-unsupervised} and mDeBERTaV3 \cite{he2023debertav3improvingdebertausing}, trained on CommonCrawl--derived corpora containing substantially more Latvian data (\textasciitilde{}1.2B tokens after cleaning). More recently, mmBERT \cite{marone2025mmbertmodernmultilingualencoder} further scales multilingual pretraining to 3T+ tokens across 1{,}800+ languages and typically outperforms XLM-R on classification, embedding, and retrieval tasks. It is designed to better support a broad range of languages by gradually adding languages during training and reducing the dominance of high-resource languages.

Several monolingual or Baltic-focused encoders have also been released. LVBERT \cite{Znotins-Barzdins:2020:BalticHLT} is trained on \textasciitilde{}0.5B tokens, primarily from news, while LitLat BERT \cite{ulvcar2021training} jointly pretrains Latvian and Lithuanian with additional English data (4.07B tokens total). The HPLT project released Latvian monolingual BERT models trained on HPLTv2, based on substantially larger cleaned Latvian text collections (3.46B tokens) \cite{samuel-etal-2023-trained,burchell2025expandedmassivemultilingualdataset}. Overall, these models show significant improvements over multilingual encoders.

Evaluation for Latvian remains fragmented and focuses mainly on syntactic tagging and NER. EuroEval\footnote{\url{https://euroeval.com/}} \cite{nielsen-2023-scandeval,saattrup-nielsen-etal-2025-encoder} recently added Latvian datasets for reading comprehension (MultiWikiQA-lv), sentiment classification (Latvian Twitter Sentiment), linguistic acceptability (ScaLA-lv), and NER (WikiANN-lv, FullStack-NER-lv) for encoder evaluation. However, there is no unified benchmark or leaderboard for Latvian encoder models.

\begin{table*}[!htbp]
\centering
\begin{tabular}{lr}
\hline
\textbf{Name} & \textbf{Word count (M)} \\
\hline
\hline
\multicolumn{2}{l}{\textbf{Web-scale and curated sources}} \\
\hline
FineWeb2 & 5371 \\
HPLT-v2 (cleaned) & 3460 \\
Crawled news & 1100 \\
Tweets & 453 \\
Books & 534 \\
Academic texts & 340 \\
Wikipedia & 53 \\
\hline
\multicolumn{2}{l}{\textbf{Latvian National Corpus Collection (LNCC)}} \\
\hline
News portal comments~\cite{rozukalne2021covid} & 642 \\
News & 513 \\
The Balanced Corpus of Modern Latvian~\cite{LVK2022} & 123 \\
Legal acts~\cite{Likumi} & 116 \\
Other & 100 \\
\hline
Total after filtering and deduplication & 6430 \\
\hline
\end{tabular}
\caption{Overview of curated Latvian text corpora. Word counts are in millions.}
\label{tab:data}
\end{table*}

\section{Pretraining}

\subsection{Dataset}
All encoder models are pretrained on the same Latvian text mixture summarized in Table~\ref{tab:data}. The corpus combines large-scale web crawls with curated Latvian resources to balance coverage, quality, and topical diversity.

We include Latvian subsets from FineWeb2~\cite{penedo2025fineweb2pipelinescale} and HPLT-v2 cleaned~\cite{burchell2025expandedmassivemultilingualdataset}. Although these datasets significantly improve multilingual coverage, Latvian content often suffers from shallow language-specific filtering, since identification and quality models are typically tuned for higher-resource languages. To improve quality without introducing heavy preprocessing, we apply additional document filtering and combine these crawls with curated Latvian collections.

The remaining sources largely come from the Latvian National Corpus Collection (LNCC)~\cite{Saulite-EtAl:2022:LREC}, including news, legal texts, comments, and other balanced materials. We further augment the mixture with a newer Latvian Wikipedia dump, newly crawled news media, tweets, academic texts, and scanned book texts. Together, these sources support both modern usage and domain variety.

All sources undergo boilerplate and low-quality text removal. We perform exact duplicate removal using metadata and text matching. Near-duplicate documents are removed using MinHash LSH with 5-grams and a similarity threshold of 0.7, following the FineWeb2 processing approach~\cite{penedo2025fineweb2pipelinescale}.

To filter low-quality or out-of-domain documents, we apply heuristics and train a 5-gram KenLM language model on the Balanced Corpus of Modern Latvian. We then score documents using perplexity: those with high perplexity are discarded, as they typically correspond to noisy, non-fluent, or weakly Latvian content.

To support context extension, we additionally sample higher quality documents with diverse token lengths, prioritizing texts longer than the original 1{,}024-token context window. The final context extension dataset contains 2.5B tokens in total: 1B tokens from documents of at least 4{,}096 tokens, 1B tokens from documents ranging between 1{,}024 and 4{,}096 tokens, and 500M tokens from shorter documents. Shorter texts are included to mitigate potential performance degradation on shorter contexts \cite{gao-etal-2025-train}.

In total, the final filtered dataset contains 6.43 billion words. The preprocessing pipeline is intentionally conservative, prioritizing precision over maximum data volume. We also highlight the need for a dedicated Latvian document quality scorer to further improve future corpus construction.

\subsection{Tokenizer}
For tokenization, we use the HPLTv2 Latvian tokenizer \cite{burchell2025expandedmassivemultilingualdataset}, a GPT-2--style byte-level WordPiece tokenizer with a vocabulary size of 32{,}768. Initial experiments did not show meaningful improvements from using a larger vocabulary.

\subsection{Architectures}
We train three model architectures. All models are pretrained on 100B tokens for fair comparison, using an effective batch size of 4.2M tokens (8k sequences of length 512) with Distributed Data Parallel (DDP). All experiments were run on a single DGX machine with 8 NVIDIA 141GB GPUs.

\paragraph{RoBERTa}
We follow the RoBERTa recipe implemented in HuggingFace Transformers \cite{liu2019roberta}. We use sequence packing and apply 30\% span-based masking \cite{joshi2020spanbertimprovingpretrainingrepresenting}.

\paragraph{DeBERTaV3}
We pretrain lv-deberta-base following the CamemBERTaV2 recipe \cite{antoun-etal-2023-data,antoun2024camembert20smarterfrench} in a single training phase using the RTD objective with a masking probability of 20\%. The generator has a hidden size of 256, and we use an effective batch size of 8k sequences per optimizer update.

\paragraph{ModernBERT}
We follow the ModernBERT training recipe \cite{modernbert,weller2025seqvsseqopen} with three-stage training. We train three model variants: mini (59M parameters), base (136M), and large (377M). We use 30\% span-based masking and apply a 2$\times$ smaller learning rate to avoid instability near the end of later stages. Training consists of: (i) 70B tokens in a stable phase with batch size 4096 and sequence length 1024, (ii) 20B tokens for context extension with sequence length 8096, and (iii) 10B tokens for cooldown using a $(1-\sqrt{\text{LR}})$ schedule. During the decay phase, masking is reduced to 15\%. This setup enables fair comparison of checkpoints before and after applying the decay phase.
ModernBERT also introduces several training optimizations, including token-level unpadding (>99\% greedy packing efficiency), FlashAttention2, and RoPE positional embeddings, yielding approximately 2$\times$ faster training.

\section{Evaluation}
We evaluate the pretrained Latvian encoder models under two complementary regimes:
(i) EuroEval-style lightweight diagnostics for quick screening under minimal-data fine-tuning, and
(ii) more in-depth evaluation on larger linguistically grounded benchmarks, including Universal Dependencies parsing and word sense disambiguation.

As baselines, we report results for existing Latvian-specific encoders and widely used multilingual pretrained models with Latvian support. Multilingual baselines include mdeberta-v3-base \cite{he2023debertav3improvingdebertausing}, xlm-roberta-large and xlm-roberta-base \cite{conneau-etal-2020-unsupervised}, mmBERT-base and mmBERT-small \cite{marone2025mmbertmodernmultilingualencoder}, as well as the original bert-base-multilingual model \cite{devlin-etal-2019-bert}. Latvian-specific baselines include hplt-bert-base-lvs \cite{burchell2025expandedmassivemultilingualdataset}, litlat-bert \cite{ulvcar2021training}, and lvbert \cite{Znotins-Barzdins:2020:BalticHLT}.


\subsection{Lightweight Tasks}
We adapt the lightweight diagnostic task suite from EuroEval \cite{nielsen-2023-scandeval}. Specifically, we use the EuroEval subsampled Latvian benchmark datasets and follow their task definitions and data preparation procedures closely. EuroEval down-samples each dataset to 1{,}024 / 256 / 2{,}048 samples for train/validation/test, and we use the same splits. In addition, we include COPA, a small commonsense reasoning benchmark, as it fits the same minimal-data and rapid fine-tuning regime.

In the original EuroEval protocol, all encoder models are fine-tuned with ten random seeds using a learning rate of $2\times10^{-5}$ and early stopping patience 2, and evaluated on bootstrapped test sets. However, we found these hyperparameters to be suboptimal for Latvian, yielding unstable model rankings. Therefore, we conducted a hyperparameter search for all models over $lr \in \{1,2,3,5,10\}\times10^{-5}$ with early stopping patience 5. For each learning rate, we trained five models with different random seeds and selected the best-performing learning rate based on the average validation performance across all metrics. We then report test-set results for the five models trained with the selected learning rate. To improve computational efficiency, we optimize padding and checkpointing, enabling the full diagnostic suite to be completed within a few GPU hours per model.

Results are reported in Table~\ref{tab:results}.

\paragraph{LTEC (Twitter Sentiment).}
A sentiment classification dataset of Latvian Twitter posts from the food and drinks domain \cite{SprogisRikters2020BalticHLT,rikters-etal-2024-annotations-exploring}. 
The original dataset contains 5{,}059 training and 754 test examples.
Inter-annotator agreement is 70.48\%.
We evaluate performance using Matthews correlation coefficient (MCC) and macro-F1 (MF1).

\paragraph{ScaLA (Linguistic Acceptability).}
A Latvian grammatical acceptability dataset derived from the Latvian UD treebank \cite{nielsen-2023-scandeval}.
Ungrammatical sentences are generated via constrained token deletion and swapping.
Results are reported with MCC and MF1.

\paragraph{FSNER (Named Entity Recognition).}
A Latvian NER dataset from a multilayer syntactic--semantic corpus (approximately 60\% news, 20\% fiction, 10\% legal, 5\% spoken, 5\% mixed) \cite{gruzitis-etal-2018-creation}.
The full dataset contains 11{,}425 samples.
We report micro-F1, both excluding the MISC class (mF1$^\dagger$) and including all classes (mF1).

\paragraph{WikiQA (Reading Comprehension).}
A multilingual Wikipedia QA dataset with LLM-generated questions and extractive answers \cite{smart2025multiwikiqareadingcomprehensionbenchmark}. 
The Latvian portion contains 5{,}000 examples.
Performance is measured using token-level F1 and exact match (EM).

\paragraph{COPA (Commonsense Reasoning).}
A Latvian translation of the English COPA dataset\footnote{\url{https://huggingface.co/datasets/AiLab-IMCS-UL/copa-lv}}, machine translated and manually post-edited \cite{skadina-etal-2025-first}. 
We further post-edit entries not corrected in the original release.
We use the original splits of 400 / 100 / 500 for train/validation/test.
We evaluate using MCC and accuracy (ACC).

\subsection{Universal Dependencies}
To assess how well pretrained encoders support Latvian morphosyntactic modeling, we fine-tune each model on Universal Dependencies (UD) tasks: part-of-speech tagging (UPOS/XPOS), lemmatization, morphological feature tagging, and dependency parsing.
We use the Latvian UD treebank from UD v2.16 \cite{nivre-etal-2020-universal}, which contains approximately 19{,}000 manually annotated sentences, and follow the official train/validation/test split.

All encoder models are evaluated within the same unified multi-task architecture. Token-level predictions are produced using shallow feed-forward classification heads with layer normalization and dropout. For dependency parsing, we employ a biaffine graph-based parser that scores head--dependent arcs and dependency relations, and we decode trees using a maximum spanning tree algorithm under a single-root constraint. We use a learning rate of $5\times10^{-4}$ for the task-specific heads and a ten-times smaller learning rate for the encoder, and report results averaged over five random seeds.

We report standard UD metrics computed with the official CoNLL UD evaluation script, including token-level tagging accuracy for UPOS, XPOS, UFeats, AllTags, and Lemmas, as well as dependency parsing quality measured by UAS, LAS, CLAS, MLAS, and BLEX.
Results are shown in Table~\ref{tab:ud}.

\subsection{Word Sense Disambiguation}
To evaluate semantic representation quality, we construct and release a manually annotated Latvian word sense disambiguation (WSD) dataset\footnote{\url{https://huggingface.co/datasets/AiLab-IMCS-UL/wsd-lv}} based on annotated example sentences from the Latvian WordNet \cite{paikens-etal-2022-towards}. These corpus examples have been manually linked to specific word senses and subsenses.

We ignore subsenses and filter out lemmas that have only a single primary sense in the inventory. The final dataset contains 1{,}821 lemma entries with 5{,}459 unique senses (approximately three senses per lemma). Each sense is associated with multiple annotated example sentences, yielding 54{,}364 labeled instances, including target word offsets.

We split the dataset by lemma entry to prevent lexical overlap between training and evaluation sets: 500 entries are reserved for testing, 200 for validation, and the remaining entries for training.

We follow the GlossBERT formulation \cite{huang-etal-2019-glossbert}, casting WSD as a context--gloss matching task. For each instance, the model scores the compatibility between the context sentence containing the target word and a candidate gloss, and selects the highest-scoring gloss at inference time. To improve discrimination among closely related senses, we sample hard negatives from competing senses within the same lemma entry.

We evaluate performance using binary accuracy on context--sense pairs and top-1 sense selection accuracy over the full primary sense inventory. We use a learning rate of $2\times10^{-5}$ and report results averaged over five random seeds. Results are shown in Table~\ref{tab:wsd}.

\begin{table}[htb]
\centering
\begin{tabular}{lrr}
\hline
\textbf{Model} & \textbf{Params} & \textbf{Vocab} \\
\hline
mdeberta-v3-base   & 278M & 250K \\
xlm-roberta-large  & 560M & 250K \\
xlm-roberta-base   & 278M & 250K \\
mmBERT-base        & 307M & 256K \\
mmBERT-small       & 140M & 256K \\
bert-base-multi    & 178M & 120K \\
\hline
hplt-bert-base-lvs & 124M & 33K  \\
litlat-bert        & 151M & 84K  \\
lvbert             & 111M & 32K  \\
\hline
lv-deberta-base       & 111M & 33K  \\
lv-mbert-large        & 377M & 33K  \\
lv-mbert-base         & 136M & 33K  \\
lv-mbert-mini         &  59M & 33K  \\
lv-roberta-base       & 124M & 33K  \\
\hline
\end{tabular}
\caption{Model size (millions of parameters) and vocabulary size (thousands).}
\label{tab:sizes}
\end{table}

\begin{table*}[htb]
\centering
\small
\setlength{\tabcolsep}{5pt} 
\begin{tabular}{l|rr|rr|rr|rr|rr}
\hline
\multicolumn{1}{l|}{\textbf{Model}}
& \multicolumn{2}{c|}{\textbf{LTEC}}
& \multicolumn{2}{c|}{\textbf{ScaLA}}
& \multicolumn{2}{c|}{\textbf{FSNER}}
& \multicolumn{2}{c|}{\textbf{WikiQA}}
& \multicolumn{2}{c}{\textbf{COPA}}
\\
\cline{2-3}\cline{4-5}\cline{6-7}\cline{8-9}\cline{10-11}
& \textbf{MCC} & \textbf{MF1}
& \textbf{MCC} & \textbf{MF1}
& \textbf{mF1$^\dagger$} & \textbf{mF1}
& \textbf{F1} & \textbf{EM}
& \textbf{MCC} & \textbf{ACC}
\\
\hline
\multicolumn{11}{l}{\textit{Existing multilingual pretrained models}} \\
\hline
mdeberta-v3-base   & 49.1\tiny±1.1 & 65.8\tiny±1.2 & 54.7\tiny±2.2 & 75.9\tiny±1.7  & 86.8\tiny±0.8 & 81.4\tiny±1.0 & 64.6\tiny±1.6 & 49.2\tiny±2.0 & \sbest{17.6}\tiny±5.5 & \sbest{58.8}\tiny±2.8 \\
xlm-roberta-large  & \sbest{52.0}\tiny±1.8 & \sbest{68.0}\tiny±1.9 & \sbest{56.0}\tiny±0.9 & \sbest{77.3}\tiny±0.4  & \sbest{87.2}\tiny±0.5 & \sbest{81.5}\tiny±0.8 & \sbest{71.6}\tiny±0.5 & \sbest{57.1}\tiny±0.7 & 1.0\tiny±4.5  & 50.5\tiny±1.7 \\
xlm-roberta-base   & 47.6\tiny±2.3 & 64.4\tiny±2.0 & 44.4\tiny±1.6 & 70.1\tiny±1.0  & 83.6\tiny±1.1 & 78.4\tiny±0.7 & 51.6\tiny±3.2 & 39.6\tiny±1.9 & 2.0\tiny±2.6  & 51.0\tiny±1.3 \\
mmBERT-base        & 45.1\tiny±1.7 & 62.8\tiny±1.7 & 44.6\tiny±0.4 & 71.7\tiny±0.7  & 80.4\tiny±1.5 & 74.8\tiny±2.2 & 64.3\tiny±1.8 & 49.6\tiny±2.2 & 16.4\tiny±2.9 & 58.2\tiny±1.4 \\
mmBERT-small       & 39.9\tiny±1.7 & 58.1\tiny±2.6 & 36.6\tiny±0.8 & 67.9\tiny±0.7  & 77.9\tiny±1.6 & 72.4\tiny±1.8 & 57.3\tiny±3.0 & 42.9\tiny±3.3 & 6.6\tiny±5.0  & 53.3\tiny±2.5 \\
bert-base-multi    & 32.2\tiny±2.5 & 53.0\tiny±3.1 & 26.6\tiny±1.0 & 61.8\tiny±1.4  & 81.0\tiny±0.7 & 75.3\tiny±0.6 & 47.0\tiny±2.7 & 35.9\tiny±2.1 & 5.5\tiny±3.8  & 52.8\tiny±1.9 \\
\hline
\multicolumn{11}{l}{\textit{Existing Latvian-specific models}} \\
\hline
hplt-bert-base-lvs & \sbest{52.5}\tiny±1.4 & \sbest{68.0}\tiny±1.0 & \sbest{56.2}\tiny±2.3 & 74.9\tiny±1.5  & \sbest{88.5}\tiny±1.3 & \sbest{83.1}\tiny±1.2 & \sbest{66.6}\tiny±1.4 & \sbest{52.7}\tiny±1.1 & 9.8\tiny±6.9  & 54.9\tiny±3.5 \\
litlat-bert        & 49.4\tiny±1.7 & 65.7\tiny±1.4 & 51.9\tiny±0.8 & 73.8\tiny±1.4  & 84.8\tiny±0.8 & 78.7\tiny±0.9 & 59.8\tiny±1.0 & 46.4\tiny±1.2 & 15.9\tiny±2.6 & 57.9\tiny±1.3 \\
lvbert             & 46.0\tiny±2.3 & 63.2\tiny±2.8 & 52.9\tiny±1.7 & \sbest{75.2}\tiny±1.3  & 85.2\tiny±0.9 & 79.1\tiny±0.6 & 24.5\tiny±3.0 & 14.8\tiny±2.3 & \sbest{17.1}\tiny±4.7 & \sbest{58.6}\tiny±2.4 \\
\hline
\multicolumn{11}{l}{\textit{This work}} \\
\hline
lv-deberta-base       & \sbest{54.2}\tiny±1.3 & \sbest{69.3}\tiny±0.8 & \best{64.2}\tiny±1.5 & \best{81.5}\tiny±1.1  & \best{89.0}\tiny±1.0 & 83.0\tiny±1.2 & 69.2\tiny±1.5 & 54.8\tiny±2.3 & \sbest{52.5}\tiny±3.5 & \sbest{76.2}\tiny±1.8 \\
lv-mbert-large        & 49.2\tiny±1.8 & 66.1\tiny±1.4 & 59.8\tiny±2.6 & 79.2\tiny±1.6  & 85.9\tiny±1.0 & 80.3\tiny±1.5 & \best{73.6}\tiny±0.3 & \best{59.6}\tiny±0.7 & 42.0\tiny±8.5 & 71.0\tiny±4.3 \\
lv-mbert-base         & 48.0\tiny±2.9 & 63.6\tiny±4.1 & 57.1\tiny±2.7 & 78.4\tiny±1.2  & 85.1\tiny±1.1 & 78.8\tiny±1.3 & 69.0\tiny±0.6 & 54.3\tiny±0.7 & 27.4\tiny±3.4 & 63.7\tiny±1.7 \\
lv-mbert-mini         & 48.8\tiny±1.2 & 65.6\tiny±1.2 & 51.3\tiny±3.4 & 74.1\tiny±2.5  & 85.1\tiny±0.8 & 79.5\tiny±0.9 & 62.8\tiny±1.3 & 48.5\tiny±1.5 & 26.7\tiny±9.1 & 63.3\tiny±4.5 \\
lv-roberta-base       & 50.7\tiny±1.8 & 66.7\tiny±1.6 & 59.0\tiny±2.3 & 78.6\tiny±1.6  & 88.7\tiny±0.6 & \best{83.6}\tiny±0.7 & 58.1\tiny±3.4 & 44.9\tiny±3.0 & 23.6\tiny±8.8 & 61.8\tiny±4.4 \\
\hline\hline
\multicolumn{11}{l}{\textit{LLM models}} \\
\hline
gpt-5  & \best{59.1}\tiny±0.7 & 71.5\tiny±0.5 & 46.5\tiny±1.1 & 72.8\tiny±0.8 & 83.0\tiny±0.6 & 72.4\tiny±0.7 & 65.1\tiny±0.9 & 46.1\tiny±1.2 & \best{96.9}\tiny±0.4 & \best{98.5}\tiny±0.3\\
gpt-5$^{\diamond}$   & 57.8\tiny±0.8 & \best{72.1}\tiny±0.5 & \sbest{57.7}\tiny±0.9 & \sbest{78.8}\tiny±0.6 & \sbest{85.6}\tiny±0.5 & \sbest{77.8}\tiny±0.6 & \sbest{65.8}\tiny±0.8 & \sbest{46.3}\tiny±1.1 & \best{96.9}\tiny±0.4 & \best{98.5}\tiny±0.3 \\
gemma-3-27b-it$^{\diamond}$      & 51.3\tiny±1.1 & 68.3\tiny±0.7 & 51.0\tiny±2.5 & 43.0\tiny±2.4 & 40.0\tiny±1.7 & 69.7\tiny±0.9 & 61.3\tiny±1.5 & 36.8\tiny±1.8 & 92.2\tiny±2.8 & 96.1\tiny±1.4 \\
\hline
\end{tabular}
\caption{Results for sentiment classification (LTEC), linguistic acceptability (ScaLA), named entity recognition (FSNER), question answering (WikiQA), and commonsense reasoning (COPA). MCC denotes Matthews correlation coefficient; MF1 macro-F1; mF1 micro-F1; and EM exact match. $^\dagger$ indicates scores excluding the MISC class. $^{\diamond}$ marks few-shot prompting for LLMs, while all other LLM results are zero-shot. The best overall result in each column is highlighted in bold, and the best result within each model group is underlined.}
\label{tab:results}
\end{table*}

\begin{table*}[ht]
\centering
\small
\setlength{\tabcolsep}{5pt} 
\begin{tabular}{lrrrrrrrrrr}
\hline
\textbf{Model} & \textbf{UPOS} & \textbf{XPOS} & \textbf{UFeats} & \textbf{AllTags} & \textbf{Lemmas} & \textbf{UAS} & \textbf{LAS} & \textbf{CLAS} & \textbf{MLAS} & \textbf{BLEX} \\
 \hline
 \multicolumn{11}{l}{\textit{Existing multilingual pretrained models}} \\
 \hline
 mdeberta-v3-base   & 98.7\tiny±0.0 & 93.8\tiny±0.1 & 97.0\tiny±0.2 & 93.5\tiny±0.2 & 97.7\tiny±0.0 & 94.4\tiny±0.3 & 92.0\tiny±0.3 & 90.5\tiny±0.2 & 86.0\tiny±0.2 & 87.9\tiny±0.2 \\
 xlm-roberta-large  & \sbest{98.8}\tiny±0.1 & \sbest{94.2}\tiny±0.2 & \sbest{97.2}\tiny±0.1 & \sbest{94.0}\tiny±0.2 & \best{98.0}\tiny±0.1 & \sbest{94.8}\tiny±0.3 & \sbest{92.5}\tiny±0.2 & \sbest{91.0}\tiny±0.3 & \sbest{86.9}\tiny±0.3 & \sbest{88.8}\tiny±0.2 \\
 xlm-roberta-base   & 98.5\tiny±0.0 & 93.5\tiny±0.1 & 96.7\tiny±0.1 & 93.3\tiny±0.0 & 97.5\tiny±0.1 & 93.8\tiny±0.2 & 91.3\tiny±0.2 & 89.6\tiny±0.2 & 85.0\tiny±0.3 & 87.0\tiny±0.1 \\
 mmBERT-base        & 98.6\tiny±0.0 & 93.7\tiny±0.0 & 96.7\tiny±0.1 & 93.4\tiny±0.0 & 97.5\tiny±0.2 & 93.8\tiny±0.2 & 91.4\tiny±0.2 & 89.8\tiny±0.2 & 85.1\tiny±0.1 & 87.1\tiny±0.4 \\
 mmBERT-small       & 98.3\tiny±0.1 & 93.1\tiny±0.2 & 96.4\tiny±0.3 & 92.8\tiny±0.3 & 97.1\tiny±0.1 & 93.2\tiny±0.3 & 90.6\tiny±0.1 & 88.8\tiny±0.1 & 83.8\tiny±0.4 & 85.7\tiny±0.1 \\
 bert-base-multi    & 98.1\tiny±0.1 & 92.4\tiny±0.2 & 95.8\tiny±0.1 & 92.1\tiny±0.1 & 96.9\tiny±0.1 & 91.5\tiny±0.3 & 88.7\tiny±0.2 & 86.8\tiny±0.2 & 81.4\tiny±0.3 & 83.9\tiny±0.2 \\
 \hline
 \multicolumn{11}{l}{\textit{Existing Latvian-specific models}} \\
\hline
hplt-bert-base-lvs & \sbest{98.9}\tiny±0.0 & \sbest{94.3}\tiny±0.2 & \sbest{97.4}\tiny±0.1 & \sbest{94.1}\tiny±0.1 & \sbest{97.6}\tiny±0.1 & \sbest{94.6}\tiny±0.2 & \sbest{92.3}\tiny±0.2 & \sbest{90.8}\tiny±0.2 & \sbest{86.8}\tiny±0.3 & \sbest{88.0}\tiny±0.4 \\
 litlat-bert        & 98.8\tiny±0.1 & 93.9\tiny±0.2 & 97.1\tiny±0.1 & 93.7\tiny±0.2 & 97.5\tiny±0.1 & 94.5\tiny±0.1 & 92.1\tiny±0.1 & 90.6\tiny±0.1 & 86.3\tiny±0.1 & 87.8\tiny±0.1 \\
 lvbert             & 98.6\tiny±0.1 & 93.2\tiny±0.1 & 96.9\tiny±0.1 & 93.0\tiny±0.1 & 96.9\tiny±0.0 & 93.6\tiny±0.2 & 91.1\tiny±0.2 & 89.6\tiny±0.2 & 85.1\tiny±0.3 & 86.2\tiny±0.1 \\
\hline
 \multicolumn{11}{l}{\textit{This work}} \\
\hline
lv-deberta-base    & \best{99.0}\tiny±0.1 & \best{94.6}\tiny±0.2 & \best{97.7}\tiny±0.0 & \best{94.4}\tiny±0.3 & \best{98.0}\tiny±0.1 & \best{95.1}\tiny±0.1 & \best{92.9}\tiny±0.1 & \best{91.4}\tiny±0.1 & \best{87.7}\tiny±0.1 & \best{89.1}\tiny±0.2 \\
 lv-mbert-large     & 98.9\tiny±0.0 & 94.4\tiny±0.0 & 97.4\tiny±0.1 & 94.2\tiny±0.0 & 97.8\tiny±0.2 & 95.0\tiny±0.4 & 92.7\tiny±0.3 & 91.1\tiny±0.4 & 87.0\tiny±0.4 & 88.6\tiny±0.5 \\
 lv-mbert-base      & 98.9\tiny±0.1 & 94.3\tiny±0.1 & 97.4\tiny±0.0 & 94.0\tiny±0.1 & 97.7\tiny±0.1 & 94.7\tiny±0.0 & 92.4\tiny±0.1 & 90.9\tiny±0.2 & 86.7\tiny±0.2 & 88.1\tiny±0.3 \\
 lv-mbert-mini      & 98.7\tiny±0.1 & 93.6\tiny±0.0 & 97.2\tiny±0.1 & 93.3\tiny±0.1 & 97.2\tiny±0.0 & 94.1\tiny±0.1 & 91.7\tiny±0.1 & 90.1\tiny±0.2 & 85.7\tiny±0.2 & 86.9\tiny±0.2 \\
 lv-roberta-base    & 98.9\tiny±0.0 & \best{94.6}\tiny±0.1 & 97.5\tiny±0.1 & \best{94.4}\tiny±0.1 & 97.9\tiny±0.0 & 95.0\tiny±0.1 & 92.7\tiny±0.0 & 91.2\tiny±0.1 & 87.4\tiny±0.1 & 88.8\tiny±0.1 \\
\hline
\end{tabular}
\caption{Universal Dependencies results on the Latvian UD treebank (UD v2.16) under joint multi-task fine-tuning. Evaluation is performed using the standard CoNLL-U scoring script.}
\label{tab:ud}
\end{table*}

\section{Results}
Tables~\ref{tab:results}--\ref{tab:wsd} summarize performance across the lightweight diagnostic tasks, UD morphosyntax, and WSD semantics. Overall, the monolingual encoders introduced in this work are competitive with multilingual baselines and prior Latvian-specific models across all evaluation regimes. In particular, lv-deberta-base achieves the most consistently strong results across tasks, despite having substantially fewer parameters than larger multilingual encoders (111M vs.\ 560M for xlm-roberta-large, i.e., $\approx5\times$ smaller; Table~\ref{tab:sizes}).

\subsection{Lightweight Tasks}
Results for the EuroEval-style diagnostic suite are presented in Table~\ref{tab:results}. Among existing models, xlm-roberta-large, mdeberta-v3-base, and hplt-bert-base-lvs establish strong baselines, frequently appearing as the top-performing models among previous work.

The lv-deberta-base model outperforms all other encoders across nearly all tasks, except for WikiQA, where it is surpassed by larger models (lv-mbert-large and xlm-roberta-large).

The ModernBERT-based models (lv-mbert) demonstrate strong competitiveness against existing multilingual encoders. However, compared to the strongest baselines, they achieve slightly weaker results on sentiment classification (LTEC) and named entity recognition (FSNER). They also generally trail lv-roberta-base, with the exception of WikiQA, where lv-mbert-large attains the best encoder performance.

ScaLA, WikiQA, and COPA demonstrate clear improvements over all baselines for most models introduced in this work, with the exception of lv-mbert-mini. Interestingly, on LTEC, only lv-deberta-base surpasses all baselines, even though the pretraining dataset contains internet comments and tweets. For FSNER, results converge across the strongest systems, with micro-F1 scores (excluding MISC) approaching 89, suggesting that performance on this task is reaching saturation across top encoders.

COPA remains particularly challenging: most multilingual and earlier Latvian models achieve only modest performance (Matthews correlation coefficient $<20$), suggesting a limited ability to discriminate between causal alternatives beyond simple heuristics. Notably, lv-deberta-base is the only encoder that attains a substantially higher score (52.5), indicating markedly stronger commonsense reasoning capability. The second-best result is achieved by lv-mbert-large.

We additionally compare encoder fine-tuning to strong commercial and open LLM baselines (Table~\ref{tab:results}).
Despite being orders of magnitude larger, LLMs do not consistently outperform the best encoder across diagnostic tasks.
One of the strongest commercial models (GPT-5) improves over lv-deberta-base on LTEC sentiment classification (MF1 71.5 vs.\ 69.3), but performs worse on ScaLA, FSNER, and WikiQA under both zero-shot and few-shot prompting.
In contrast, LLMs dominate the COPA reasoning task, achieving near-perfect performance (e.g., MCC 96.9), whereas even the best encoder remains substantially lower.

\subsection{Universal Dependencies}
UD results are reported in Table~\ref{tab:ud}. On token-level tagging tasks (UPOS, XPOS, UFeats, Lemmas), the best-performing models largely saturate performance, reflecting the relatively large size and high quality of the Latvian UD treebank. As a result, differences between encoders are small for these local classification objectives.

In contrast, larger gaps emerge for the parsing-oriented metrics (LAS, CLAS, MLAS, BLEX), which better capture structural prediction quality. While the overall differences among top-performing encoders remain modest, and no model shows a decisive advantage across all metrics, lv-deberta-base achieves the strongest and most consistent results on average, improving over both multilingual baselines and prior Latvian-specific models. These gains suggest that monolingual pretraining contributes most strongly to syntactic structure prediction beyond token-level tagging.

\begin{table}[ht]
\centering
\begin{tabular}{lrr}
\hline
 \textbf{Model}               & \textbf{Match Acc.} & \textbf{Sense Acc.}  \\
 \hline
 \multicolumn{3}{l}{\textit{Existing multilingual pretrained models}} \\
 \hline
 mdeberta-v3-base   & 79.6\tiny±0.6      & 73.5\tiny±0.7      \\
 xlm-roberta-large  & \sbest{80.9}\tiny±1.1      & \sbest{76.9}\tiny±0.5      \\
 xlm-roberta-base   & 75.3\tiny±0.6      & 65.4\tiny±0.6      \\
 mmBERT-base        & 76.9\tiny±0.8      & 69.3\tiny±1.0      \\
 mmBERT-small       & 74.1\tiny±1.3      & 62.1\tiny±0.5      \\
 bert-base-multi    & 71.3\tiny±0.9      & 56.5\tiny±0.5      \\
 \hline
 \multicolumn{3}{l}{\textit{Existing Latvian-specific models}} \\
\hline
 hplt-bert-base-lvs & \sbest{77.4}\tiny±0.5      & \sbest{72.8}\tiny±0.2      \\
 litlat-bert        & 76.2\tiny±0.8      & 67.3\tiny±0.6      \\
 lvbert             & 76.5\tiny±0.7      & 68.1\tiny±0.4      \\
 \hline
 \multicolumn{3}{l}{\textit{This work}} \\
\hline
 lv-deberta-base    & \best{83.6}\tiny±0.2      & \best{78.9}\tiny±0.3      \\
 lv-mbert-large     & \sbest{81.7}\tiny±1.2      & \sbest{77.6}\tiny±0.4      \\
 lv-mbert-base      & 80.1\tiny±0.7      & 73.8\tiny±0.5      \\
 lv-mbert-mini      & 76.3\tiny±0.5      & 67.6\tiny±0.7      \\
 lv-roberta-base    & 78.9\tiny±0.6      & 71.4\tiny±0.6      \\
\hline
\end{tabular}
\caption{WSD results. Match Acc.\ denotes binary accuracy for context--sense pairs, and Sense Acc.\ denotes top-1 sense selection accuracy.}
\label{tab:wsd}
\end{table}

\subsection{Word Sense Disambiguation}
WSD results are shown in Table~\ref{tab:wsd}. lv-deberta-base achieves the highest accuracy (78.9\% sense selection accuracy and 83.6\% binary context--sense matching accuracy), with consistent improvements over both multilingual models and existing Latvian encoders. lv-mbert models show clear gains with increased model size and outperform similarly sized baselines, performing comparably to lv-roberta-base.
Overall, the dataset provides good separation between model performances while exhibiting relatively low variance across runs, making it a reliable and informative evaluation task.

\section{Conclusion}
We introduced a suite of Latvian pretrained encoder models based on RoBERTa, DeBERTaV3, and ModernBERT architectures, including long-context variants supporting up to 8{,}192 tokens, and compiled a comprehensive set of benchmark tasks for systematic model evaluation.

Across all benchmark tasks, lv-deberta-base achieves the strongest overall performance despite its relatively small size (111M parameters). Among the diagnostic tasks, the largest relative improvement is observed in commonsense reasoning: on COPA, lv-deberta-base reaches a Matthews correlation coefficient of 50.9, whereas most multilingual and prior Latvian encoders yield only modest correlations (MCC $<20$). On the newly introduced WSD benchmark, lv-deberta-base attains the best sense selection accuracy (78.9\%) and the highest context--sense matching accuracy (83.6\%).

ModernBERT models provide competitive performance while enabling substantially higher throughput and long-context processing (up to 8{,}192 tokens). Long-context variants generally match their short-context counterparts and yield the largest gains on extractive question answering. However, despite these practical advantages, ModernBERT models remain consistently below lv-deberta-base while achieving results comparable to or slightly lower than lv-roberta-base. Increasing model size from mini to large leads to improvements across most tasks.

Overall, our results confirm that Latvian-focused monolingual pretraining yields substantial gains over multilingual encoders across both lightweight diagnostic tasks and linguistically grounded evaluations. We release all pretrained models and evaluation resources to support reproducible Latvian NLP research and facilitate downstream applications.

\section*{Limitations}
All models in this work are trained only on Latvian data. This monolingual scope limits cross-lingual transfer and may reduce robustness on mixed-language inputs, code-switching, and multilingual settings where multilingual encoders can be advantageous.

Although the training corpus is compiled from multiple sources, it inherits biases and noise typical of web- and news-derived text. The distribution is likely skewed towards formal written registers, while conversational language, dialectal variation, and minority varieties of Latvian are underrepresented. Some documents may also contain duplicated text, boilerplate, low-quality passages, or factual errors, which can affect the resulting representations.

Our evaluation focuses on task-oriented NLU benchmarks, UD morphosyntactic modeling, and supervised word sense disambiguation. The results do not directly measure retrieval and embedding quality, robustness under domain shift, performance on specialized domains, or behavior on long-context tasks beyond those included in our benchmark suite.

\section*{Acknowledgments}

This work was funded by the EU Recovery and Resilience Facility project ``Language Technology Initiative'' (2.3.1.1.i.0/1/22/I/CFLA/002).

\bibliography{custom}

\appendix


\end{document}